\documentclass[letterpaper, 10 pt, conference]{ieeeconf}  % Comment this line out if you need a4paper
\IEEEoverridecommandlockouts                              % This command is only needed if 
\usepackage{amsmath} % assumes amsmath package installed
\usepackage{amssymb}  % assumes amsmath package installed

\usepackage{balance} % makes columns on last page balanced
\usepackage{mathrsfs} % mathscr font
\usepackage{dsfont} % mathds font
\usepackage{siunitx} % For formatting of SI units: \si
\usepackage{color} % For Inkscape pdf_tex files

\usepackage{hyperref}
\usepackage[pdftex]{graphicx}
\graphicspath{{./figures/}}
\DeclareGraphicsExtensions{.pdf,.jpeg,.jpg,.png}

\usepackage{algorithm}% http://ctan.org/pkg/algorithms
\usepackage{algpseudocode}% http://ctan.org/pkg/algorithmicx

\usepackage[cal=boondox,scr=boondoxo]{mathalfa} % lower case mathscr
\usepackage{letltxmacro}
\LetLtxMacro{\vec}{\vector}
\renewcommand{\vec}[1]{\mathbf{#1}}
\newcommand{\mat}[1]{\boldsymbol{#1}}
\newcommand{\set}[1]{\mathscr{#1}}
\newcommand{\norm}[2]{\left\Vert#1\right\Vert_#2}

\DeclareMathOperator*{\argmin}{arg\,min}

\newcommand{\realset}{\mathds{R}}

\newcommand{\transpose}[1]{#1^\top}

\newcommand{\bigOh}[1]{\mathcal{O}\left(#1\right)}

\usepackage{amsthm}
\newtheorem{problem}{Problem}

\theoremstyle{definition}

\usepackage{cite}
\makeatletter
\let\MYcaption\@makecaption
\makeatother

\usepackage[font=footnotesize]{subcaption}

\makeatletter
\let\@makecaption\MYcaption
\makeatother

\newcommand{\preSubCaptionSpace}[0]{\vspace{-1.75em}}
\newcommand{\postSubCaptionSpace}[0]{\vspace{0.0em}}
\newcommand{\preCaptionSpace}[0]{\vspace{-1.0em}}
\newcommand{\postCaptionSpace}[0]{\vspace{-1.25em}}

\usepackage[usenames,dvipsnames]{xcolor}

\title{\LARGE \bf
    Estimation of Spatially-Correlated Ocean Currents from\\ Ensemble Forecasts and Online Measurements
}

\author{K.~Y.~Cadmus~To$^1$,
        Felix~H.~Kong$^1$,
        Ki~Myung~Brian~Lee$^1$,
        Chanyeol~Yoo$^1$,
        Stuart Anstee$^2$,
        and Robert~Fitch$^1$% <-this % stops a space
    \thanks{This research is supported by an Australian Government Research Training Program (RTP) Scholarship, Australia's Defence Science and Technology Group, Australian Bureau of Meteorology, and the University of Technology Sydney.}
    \thanks{$^1$Authors are with the University of Technology Sydney, NSW 2007, Australia {\tt\footnotesize \{Cadmus.To,Brian.Lee\}@student.uts.edu.au} and {\tt\footnotesize \{Chanyeol.Yoo,Felix.Kong,Robert.Fitch\}@uts.edu.au}}
    \thanks{$^2$Author is with the Defence Science and Technology Group, Department of Defence, Australia {\tt\footnotesize stuart.anstee@dst.defence.gov.au}}
}

\begin{document}
\thispagestyle{empty}
\pagestyle{empty}

%-------------------------------------------------------------
% Title
%-------------------------------------------------------------
\IEEEauthorblockconfadjspace{-1em} % CT: This command is only available by changing ieeeconf.cls !!!
\IEEEaftertitletext{\vspace{-1em}}

\maketitle

%-------------------------------------------------------------
% Abstract
%-------------------------------------------------------------
\begin{abstract}
We present a method to estimate two-dimensional, time-invariant oceanic flow fields based on data from both ensemble forecasts and online measurements.
Our method produces a realistic estimate in a computationally efficient manner suitable for use in marine robotics for path planning and related applications.
We use kernel methods and singular value decomposition to find a compact model of the ensemble data that is represented as a linear combination of basis flow fields and that preserves the spatial correlations present in the data.
Online measurements of ocean current, taken for example by marine robots, can then be incorporated using recursive Bayesian estimation.
We provide computational analysis, performance comparisons with related methods, and demonstration with real-world ensemble data to show the computational efficiency and validity of our method.
Possible applications in addition to path planning include active perception for model improvement through deliberate choice of measurement locations.
\end{abstract}

%-------------------------------------------------------------
% Introduction
%-------------------------------------------------------------
\section{Introduction}
% vision
% estimate ocean current critical for robotics
% all mobility depends on it, yet predications not available in form of resolution and timeliness sufficient for path planning
% we are interested in exploiting the ability of robots to take measurements as a way to augment predictions and produce spatially-coherent estimate in probabilistic form
% real-time, high resolution estimates are useful for robotics, and potentially also to marine science and related disciplines
Estimates of ocean current are critical for many applications of marine robots, particularly in supporting autonomous mobility. 
% However, available ocean forecast data lacks sufficient spatial resolution and certainty to be suitable for planning. 
However, currently available ocean forecast data is in a form that is not directly compatible with existing planning algorithms, and is difficult to update using online sensor measurements.
We are interested in exploiting robots' ability to take measurements of ocean current online as a way to augment forecast data in order to produce realistic flow field estimates in probabilistic form. Real-time estimates are relevant to marine robotics, and potentially also to marine science and related disciplines.

% \begin{itemize}
%     \item Ensemble predictions consists of largely similar flow patterns with some variation between them them
%     \item Estimate a flow field using only combinations of  the fundamental flow patterns that exist in the data, as opposed to estimating every flow vector across space
%     \item Required properties: Online, low bandwidth?
%     \item Low computation for onboard computation
% \end{itemize}

% motivation
% \begin{itemize}
%     \item Ocean monitoring is useful
%     \begin{itemize}
%         \item Oceanography
%         \item Military
%         \item Mining
%     \end{itemize}
% \end{itemize}

\begin{figure}[tb]
    \centering
    \includegraphics[width=0.9\columnwidth]{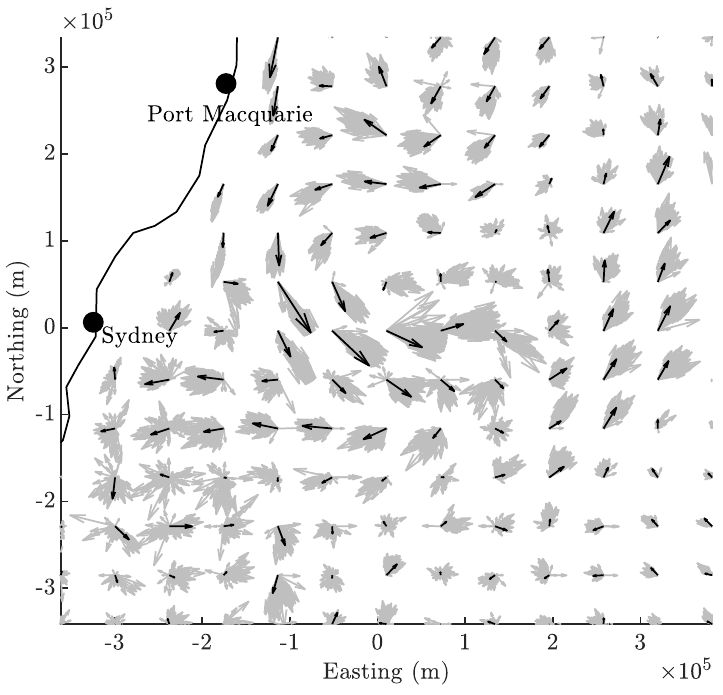}
    \preCaptionSpace
    \caption{
        Subsampled ensemble flow field data from the Australian Bureau of Meteorology~(BOM) off the coast of New South Wales, Australia.
        This ensemble forecast consists of ${N_E=96}$ flow fields, each predicting ocean surface conditions for 16$^\text{th}$ November 2018.
        One of the 96 flow fields is shown in black, whilst others are shown in grey.
        % Notice there are regions where predicted velocity vectors among the ensemble members are all relatively similar, and regions where predictions are more dispersed.
        % The distance between Sydney and Port Macquarie is approximately~$314$\,\si{km}.
    }
    \postCaptionSpace
    \label{fig:bom_ensemble}
\end{figure}

% ensemble forecasts
% \item We get multiple predictions of an ocean flow field might be (from ensemble methods), but none of them is likely to exactly correct
% however, have largely similar flow patterns
% interesting to consider combinations of these as opposed to individual vectors
% online measurements can help
Forecast data from organisations such as the Australian Bureau of Meteorology~(BOM) are often produced using ensemble forecasting, where a set (or, \emph{ensemble}) of predicted flow fields is generated from a range of initial conditions. 
While none of the ensemble members themselves are likely to be exactly correct, the ensemble as a whole tends to contain instances of largely similar flow patterns~(Fig.~\ref{fig:bom_ensemble}).
% However, none of the ensemble members themselves are likely to be exactly correct over their entire spatial extent.
% % Each member of the ensemble is considered to be equally likely, and none are likely to be exactly correct over their entire spatial extent. 
% Nevertheless, the ensemble as a whole tends to contain instances of largely similar flow patterns~(Fig.~\ref{fig:bom_ensemble}). 
Measurements taken by sensing systems could help to reduce uncertainty by observing true flow conditions locally.

%%%%%%%%%%%
% challenge
%
% Data has high dimensionality with complex interactions with its intrinsic variables (temperature, salinity, etc)
% High computation demand from high dimensionality, which is prohibitive for oceanic robotics that require online planning (cite CY's ICRA2021, CT's ICRA2020)
% also probabilistic, which is good property for use in existing robot algorithms for planning under uncertainty (typically probabilistic) but ensemble members are equally likely

%Unfortunately, flow fields have many degrees of freedom.
The ensemble format is awkward for robotics applications because planning algorithms typically assume a single probabilistic environment model, as opposed to a set of predictions~\cite{Wang2016,Kularatne2018ICRA}. 
Ensemble forecasting models flow fields by incorporating information including temperature, salinity, and the application of Navier-Stokes equations.
The challenge in producing a single model is how to distil the information contained in the ensemble in a way that captures uncertainty and that preserves spatial correlations seen in the ensemble data.

% RF this argument didn't work because couldn't find a crisp statement of why we can't use GPs naively. Possibly add back in to revision if we can find something to say...
%Gaussian process~(GP) regression is commonly employed in robotics to address similar estimation problems and can incorporate additional observations online as we propose, but unfortunately cannot be naively applied here because...

%However these are computationally prohibitive for real-time applications such as planning~\cite{Cadmus2020,Chanyeol2021}.
% Not sure how to integrate probability here
% computational part is re the planner; how to distil many ensembles into a single estimate with uncertainty
% extension of a GP regression model

%%%%%%%%%%%%
% approach
%
% Express ensemble members as continuous flow fields with the incompressible kernel and compress the representation with SVD resulting in continuous representations of basis flow fields
% Using an estimate of the flow field from the ensemble data, the algorithm refines its estimate from direct flow velocity measurements in the environment

% fk
% - ASSUME the prev paragraph just said "how do we get a single probabilistic model from an ensemble?"
% - Here's a novel method to estimate this single probabilistic model.
% - We use a novel combo of kernel methods and SVD to represent the space of possible flow fields in a way that enforce spatial correlations found in the ensemble data. 
% - Then we use a recursive Bayesian estimator to match online measurements in a ``lightweight'' way. 
In this paper, we present a novel estimation method that produces a single probabilistic model using kernel methods and singular value decomposition~(SVD). Intuitively, we find a set of basis flow fields and identify a compressed model, represented as a linear combination of these basis flows, that enforces spatial correlations found in the ensemble data. This compact representation is key to achieving computational efficiency. Then, we use a recursive Bayesian estimator to efficiently integrate online measurements. 

% RF again, didn't quite work but perhaps in a revision...
%Usage of the resulting estimator in practical applications is similar to that of GP regression.

% In this paper, we reduce the search space of the estimated flow field by enforcing the incompressibility constraint, and through a novel interpretation of ensemble forecasts.
% Recurring flow patterns, present in various quantities across all the predicted flow fields, are extracted from the data and the linear combination of these  is used to represent the estimated flow field.
% We propose to estimate the weights of the flow patterns in a Kalman filter~\cite{Kalman1960}, which is updated online with flow velocity measurements.
% search space means space of all flow fields
% needs thinking
% searching over flow fields that are in the span of the original ensemble dataset
% this helps preserve the spatial correlation
% methods we use are kernel, svd, ...

%%%%%%%%%%%%%%%%%%%
% results summary
% Our paper proposes a novel method to estimate a 2D, time-invariant oceanic flow field by constructing an observer on a low-dimensional latent state with linear ``dynamics''.}
% Experiments compared against incompressible GP and kernel observer highlights its computational advantage
% Constant iteration time complexity wrt number of measurements
% Furthermore, the algorithm allows us to take ``local measurements for global updates''
% Exploration with different measurement patterns suggests further research in coordinated sampling methods
% Promising framework for estimation of time-variant flow fields

We present the details of our method in the case of two-dimensional, time-invariant flows and provide computational analysis.
Performance comparisons with the incompressible Gaussian process~(GP) and kernel observer methods highlight the computational advantages of our method empirically.
Further, we demonstrate the behaviour of our method using ensemble data from the Australian BOM.
The results of this experiment suggest applications of our method in an active perception context where measurement locations are intentionally chosen to reduce model uncertainty.

% We provide theoretical analysis for the computational advantages of our approach compared to existing work and demonstrate the relative performance between these techniques empirically.
% Furthermore, we show the effects of using different measurement policies on real ensemble data~(Fig.~\ref{fig:bom_ensemble}) stretching across hundreds of kilometres from the Australian BOM.
% The results of this experiment suggests... \ct{CY suggested something for here}
% % active perception to guide where to take measurements

%-------------------------------------------------------------
% Related Work
%-------------------------------------------------------------
\section{Related work}

% It joins and extends upon an existing corpus of work in the literature that examine the use of data-based methods for estimating the evolution of complex spatiotemporal phenomena.
% The proposed method in this paper brings together kernel methods, the singular value decomposition~(SVD), and observers.

The problem of oceanic flow field estimation is certainly not new. 
Indeed, while an ensemble forecast is considered an input to our problem, the ensemble forecast itself~(e.g.~\cite{Leutbecher2008,Krishnamurti2000}) is used to aid human meteorological forecasters in estimating atmospheric and oceanic conditions. 
% For the purpose of robotic navigation, however, the direct generation of ensembles requires solving nonlinear partial differential equations, which is too computationally intensive for most field robotics applications.
Since an ensemble comprises many estimates, the use of filters has been previously proposed to give a single estimate of the most likely flow field. The ensemble Kalman filter~\cite{Houtekamer2001,Evensen2003} is an extension of the celebrated Kalman filter~\cite{Kalman1960} which uses the sample covariance of ensemble forecasts; additionally, particle filtering methods which do not depend on Gaussian assumptions have also been proposed for use with ensembles~\cite{Moradkhani2005,Holm2020}. However, direct implementation of these filters often requires copious amounts of computation power, due to their high dimensionality.

Another tool used for oceanic flow field estimation is the Gaussian process~(GP), also known as ``kriging''. 
A special GP formulation for oceanic flow field estimation has been proposed that imposes an incompressibility constraint, giving it faster convergence compared to a standard radial basis function kernel~\cite{Brian2019}.
%The use of a GP in conjunction with a Kalman Filter to estimate complex spatiotemporal phenomena, such as oceanic flow, has also been proposed~\cite{Mardia1998}. 
% However, the use of GPs as proposed by these methods do not take advantage of the spatial correlation between an ensemble's member flow fields, and hence require a good measurement coverage of an area to arrive at an accurate flow field estimate.
% \fk{Talk about incompressible GP}
% This incompressible kernel was originally proposed for use with a Gaussian process~(GP) for the purpose of estimating oceanic flow fields~\cite{Brian2019}.
% The current paper extends upon this work through its use of the SVD and the application of an observer on the low-dimensional latent state.
% These improvements, as we will show in this paper, allow faster convergence \fk{, better error????}, and reduced computational complexity in the number of measurements. 
A closely related class of approaches is the kernel observer~\cite{Mardia1998,Kingravi2016,Whitman2017}, which constructs an observer on the latent state of a kernel model. The use of these methods on an ensemble forecast has not been studied, however; typically kernel observers make use of a single dataset. Our method can be considered an extension of the kernel observer method which makes use of ensemble forecasts as a valuable prior, resulting in a lower-dimensional latent state and faster convergence rates.

% The authors propose and analyse the use of a kernel method to identify the latent state, and apply an observer directly to this ``kernel latent state''.
% While there are similarities, our approach extends upon this idea of kernel observers in several ways.
% Firstly, we make use of a special kernel which enforces incompressibility in the estimated flow field~\cite{Brian2019}, which has been shown to improve convergence rate in oceanic flow field estimation.
% Secondly, our approach considers a different observer state.
% In our approach, a lower-dimensional latent state is extracted from the ``kernel latent state'' via the SVD, which is used as the observer state.
% This smaller observer state confers computational advantages while preserving the linearity of the observer dynamics, even in the time-varying case not considered here. 

Another related family of methods is based on the dynamic mode decomposition~(DMD), which also has not been applied for ensemble data.
DMD-based methods also seek to obtain a low-dimensional latent state with linear dynamics, and have been used successfully to represent dynamics of periodic fluid flows~(e.g.~\cite{Kutz2016}).
Furthermore, the use of an observer on the ``DMD latent state'' has been investigated for fluid flow estimation~\cite{Tu2013,Nonomura2018}, and other time-series data~\cite{Fathi2020}, but not ensemble data.
Our approach extends upon ``DMD observer'' idea by using a kernel embedding to enforce incompressibility, which is a common model for ocean currents~(e.g. \cite{Kowalik1993,Madec2008}), and has been shown to improve error and convergence rate of a GP in real-world 2D oceanic flow fields \cite{Brian2019}.

In summary, the approach proposed in this paper can be seen as a novel combination of kernel methods, the SVD, and an observer for use with ensemble forecasts. It can be seen as a unification of the kernel observer with the ``DMD observer''. Additionally, the proposed method leverages an ensemble forecast as a prior, leading to fast convergence to accurate estimates compared to these existing methods.

% \begin{itemize}
%     \item Ensemble methods
%     \begin{itemize}
%         \item Meteorology: weather forecasting
%         % \item Why we can't use these: None of them are expected to be exact
%         \item Accurate modelling is computationally prohibitive for online applications
%         \item They act as approximate samples in an unknown complex distribution for flow field of interest \fk{Check particle filtering literature for language}
%         \item \bl{also look at ensemble Kalman filter}
%     \end{itemize}
%     \item DMD and KF for denoising time-series data\cite{Fathi2020}
%     \item DMD and KF for field of fluids~\cite{Tu2013}
%     \begin{itemize}
%         \item They don't use kernel embedding/kriging though
%         \item Uses a specific observation matrix based on the model they physically test on \fk{Does this mean $\mat H(\vec x)$ is fixed?} \ct{No I \emph{think} it's still a function of $\vec{x}$}
%         \item Doesn't use ensemble, but a series of data across time
%     \end{itemize}
%     \item DMD and KF \cite{Nonomura2018}
%     \begin{itemize}
%         \item Assumes measurements of flow fields at fixed points in space
%     \end{itemize}
%     % \item In kernel PCA, the mapping from latent variables (weights) to observables is nonlinear, but you use kernel trick to get around it. 
% \end{itemize}

%-------------------------------------------------------------
% Problem formulation
%-------------------------------------------------------------
\section{Problem formulation and approach}

We consider time-invariant \emph{ensemble forecasts} on a compact subset of two-dimensional space~$\realset^2$.
An ensemble forecast~$\set{E}$, common in meteorological weather prediction, is a set of $N_E$~\emph{ensemble members}~${\set{e}_i \in \set{E}}$.
% Each member is a deterministic flow field describing the velocity vector at a discrete set of $N_V$ locations~${\set{X}_\text{ens} \subset \realset^2}$.
Each of the ensemble members~$\set{e}_i$ is deterministically generated using slightly different initial conditions which result in different sets of \emph{flow vectors}, representing multiple predictions of the ocean current velocities at a discrete set of $N_V$~locations~${\set{X}_\text{ens} \subset \realset^2}$.
Each ensemble member is such that~$\set{e}_i = \{ \vec{z}_{i,1}, \cdots, \vec{z}_{i,N_V} \}$ where~${\vec{z}_{i,j} = \transpose{[u_{i,j}, v_{i,j}]}}$ is a flow vector for $i$-th ensemble member at position~${\vec{x}_j \in \set{X}_\text{ens}}$.
% The set of all flow vectors is denoted as~$\Lambda = \{\vec{z}_{i,j}\}$ for all ensemble members~$\set{e}_i$ and ensemble positions~$\vec{x}_j$.

% However, none of the ensemble members can be depended upon to be equal to the true, unknown flow field, as they are merely predictions.
% So in our problem, a sequence of point measurements of the true flow field are also made.
% A set of $N_K$ measurements~$\set{Z}$ of the true flow field are taken at a set of locations~$\set{X}_\text{mea}$.
% Each measurement~$\tilde{\vec{z}}\in\set{Z}$ is a measurement of the flow vector at~$\vec{x}_k\in\set{X}_\text{mea}$ that is subject to Gaussian noise:
A set of $N_K$~measurements, denoted as~$\set{Z}$, is taken at a set of positions~$\set{X}_\text{mea}$. While the positions are fully-known, each measurement~$\tilde{\vec{z}}_k\in\set{Z}$ at~$\vec{x}_k\in\set{X}_\text{mea}$ is subject to Gaussian noise, such that
\begin{equation}
    \tilde{\vec{z}}_k = \vec{f}_\text{tru}(\vec{x}_k) + \vec{n}_k
    ,
\end{equation}
where~$\vec{f}_\text{tru}:\realset^2\to\realset^2$ is the true, continuous, unknown flow field and~${\vec{n}_k \sim \mathcal{N}(\vec{0}_2,\mat{\Sigma}_\text{mea}})$ is the i.i.d. measurement error with covariance~$\mat{\Sigma}_\text{mea}\in\realset^{2\times 2}$.
% Note that measurement position~$\vec{x}=\transpose{[x,y]}$ is known without noise and measurement noise is independent of position.

% Given the ensemble forecast $\set{E}$ and the set of measurements $\set{Z}$, we aim to produce a continuous flow field~${\hat{\vec{f}}:\realset^2\to\realset^2}$ that closely approximates the true flow field $\vec{f}_\text{tru}$ everywhere on the domain. 
% Ideally, we want to find a smooth flow field estimate~$\hat{\vec{f}}$ that minimises
% % \begin{equation}
% %     \label{eqn:performanceMetric}
% %     J(\hat{\mathbf f}) = \sum_{\mathbf x\in\mathbf X_\text{test}} |\mathbf f_\text{tru}(\mathbf x) - \hat{\mathbf f}(\mathbf x)|^2,
% % \end{equation}
% % where $\mathbf X_\text{test}$ is a set of test locations with which to evaluate the accuracy of $\hat{\mathbf f}$. 
% \begin{equation}
%     \label{eqn:performanceMetric}
%     J(\hat{\mathbf f}) = \norm{\vec{f}_\text{tru} - \hat{\vec{f}}}{2}^2,
% \end{equation}
% where $\norm{\cdot}{2}$ denotes the $\mathcal{L}^2$ norm of a function.
% However, in practical situations, obviously $\vec{f}_\text{tru}$ is unknown, and cannot be directly evaluated.
% Since the only available data about $\vec{f}_\text{tru}$ enters the problem via the measurements $\set{Z}$, we instead formulate the problem in the following way.

We aim to estimate the continuous flow field~${\hat{\vec{f}}:\realset^2\to\realset^2}$ given the ensemble forecast~$\mathcal{E}$ and a set of measurements~$\set{Z}$. 
% Formally,
\begin{problem}
\label{prob:fitting}
    Given a finite set of $N_K$ noisy measurements~$\tilde{\vec{z}}_k\in\set{Z}$, at locations $\vec{x}_k$, find a continuous flow field $\hat{\vec{f}}$ that minimises
    \begin{equation} \label{eqn:fitToMeasurements}
        \hat{\vec{f}}^* = \argmin_{\hat{\vec{f}}\in C^\infty(\vec{x})}\sum_{k=1}^{N_K}\norm{\tilde{\vec{z}}_k - \hat{\vec{f}}(\vec{x}_k)}{2}^2
        .
    \end{equation}
\end{problem}
\noindent However, \eqref{eqn:fitToMeasurements} is severely underdetermined; additional properties must be imposed to constrain the problem.
The main desirable property of $\mat{\vec{f}}^*$ is to preserve spatial correlation (i.e. flow patterns) within the ensemble~$\set{E}$. 

\subsection{Approach overview} \label{sec:overview}
Our proposed approach solves Prob.~\ref{prob:fitting} with the property that $\hat{\mathbf{f}}^*$ preserves the spatial correlations in $\set E$.
% % As stated previously, \eqref{eqn:fitToMeasurements} is underdetermined.
% A standard method for improving ill-posed problems such as~\eqref{eqn:fitToMeasurements} is Tikhonov regularisation.
% % \begin{equation}
% %     \hat{\vec{f}}^* = \argmin_{\hat{\vec{f}}\in C^\infty(\vec{x})}\sum_{k=1}^{N_K}\left(\norm{\tilde{\vec{z}}_k - \hat{\vec{f}}(\vec{x}_k)}{2}^2\right) + \lambda \norm{\hat{\vec{f}}}{2}^2
% %     ,
% % \end{equation}
% % for some $\lambda > 0$.
% For the regularised problem, the representer theorem~(e.g. \cite{Rasmussen2006}) asserts that the optimal solution~$\hat{\vec{f}}^*$ can be expressed in terms of a finite-dimensional basis~${\mat{H}:\realset^2\to\realset^{2\times N_W}}$ and weight vector~${\vec{w}\in \realset^{N_W}}$:
% \begin{equation} \label{eqn:flowPrediction_weight}
%     \hat{\vec{f}}^*(\vec{x}) = \mat{H}(\vec{x})\vec{w}
%     ,
% \end{equation}
% where $\vec{x}=\transpose{[x,y]}$ is a location.
To do this, we propose to construct $\hat{\vec f}^*(\vec x)$ using a basis
${\mat{H}:\realset^2\to\realset^{2\times N_W}}$ and weight vector~${\vec{w}\in \realset^{N_W}}$:
\begin{equation} \label{eqn:flowPrediction_weight}
    \hat{\vec{f}}^*(\vec{x}) = \mat{H}(\vec{x})\vec{w}
    ,
\end{equation}
where $\vec{x}=\transpose{[x,y]}$ is a location.

% But how should $\mat{H}(\vec{x})$ and $\vec w$ be chosen?
% This motivates a sub problem of Prob.~\ref{prob:fitting}, which we call in this paper the problem of \emph{representation}.
% The main desirable property of $\mat{H}(\vec{x})$ is the preservation of spatial correlation, or ``flow patterns'', within the ensemble $\set{E}$.
Our approach is to choose $\mat{H}(\vec{x})$ based on the ensemble data~$\set{E}$, and to choose $\vec{w}$ based on online measurements~$\set{Z}$.
Choosing $\mat{H}(\vec{x})$ in this way ensures that the ``flow patterns''~(i.e. spatial correlations) of the resulting $\hat{\vec{f}}$ are consistent with the ensemble data $\set{E}$. 
Additionally, this constrains the valid solutions to be in the span of $\mat H(\vec x)$, excluding many spurious solutions allowed by~\eqref{eqn:fitToMeasurements}, and hence aiding the ill-posedness of the problem.
% Additionally, our choice of $\mat H(\vec x)$ includes a kernel matrix, allowing the \textit{representer theorem} to assert that  the optimal solution to the Tikhonov regularization of \eqref{eqn:fitToMeasurements} can be expressed as \eqref{eqn:flowPrediction_weight}.

% \fk{\emph{The following paragraph probably belongs in approach:} To address the problem of representation, a judicious choice of basis will be required. To do this, we interpret the ensemble data $\mathbf E$ as a collection of \emph{likely} possible flow vectors, and choose our representation to be within the span of these flow fields. This significantly restricts the search space from the space of all possible flow fields to the space of all flow fields spanned by the ensemble $\mathbf E$.}

We propose a two-stage estimation framework consisting of offline and online components.
In our approach, since $\mat{H}(\vec{x})$ does not depend on $\set{Z}$, it can be computed offline.
The basis~$\mat{H}(\vec{x})$ is chosen based on the ensemble forecast~$\set{E}$ and an additional incompressibility prior.
The online component uses a recursive filter to iteratively update $\vec{w}$ since the measurements are often sequential.
% Fig.~\ref{fig:overview} summarizes the entire framework.
% - Mention an example will be used along the approach to aid understanding

% \begin{figure}[tb]
%     \centering
%     \includegraphics[width=\columnwidth]{overview}
%     \caption{
%         Overview of our approach for flow field estimation with ensemble data and a sequence of measurements
%     }
%     % TODO:
%     % - Possibly example visuals
%     \label{fig:overview}
% \end{figure}

%-------------------------------------------------------------
% Approach: Offline section
%-------------------------------------------------------------
\section{Flow field representation} \label{sec:offline}
In this section, we describe the offline part of the proposed algorithm, comprising two steps: ``regression" and ``compression".
Firstly, we model each of the $N_E$ ensemble members resulting in continuous and incompressible representations of the ensemble flow fields encoded in latent states.
% Then, we perform an SVD on these latent states, resulting in a low-dimensional representation of the estimated latent state,
% Then, we extract the spatial correlations in $\set{E}$ via the SVD, resulting in an~$\mat{H(\vec{x})}$ that enables a parsimonious representation of any flow field in the span of~$\set{E}$. \fk{????}
% Then, we extract the spatial correlations in $\set{E}$ via the SVD, resulting in an~$\mat{H(\vec{x})}$ that enables a parsimonious representation of any set of flow vectors at~$\set{X}_\text{ens}$ in the span of~$\set{E}$. 
Then, we extract the flow patterns from $\set{E}$ with the SVD to construct a basis $\mat{H}(\vec{x})$ that preserves spatial correlations in $\set{E}$.

\subsection{Flow field regression via kernel embedding}
% \fk{We can reduce the scope of the problem by constraining search space to consider only the subspace of incompressible functions $\hat{\vec{f}}$.
% At the mesoscale, ocean currents are often modelled as incompressible~(e.g.~\cite{Kowalik1993,Madec2008}), suggesting that true oceanic flow fields are close to incompressible, and that by asserting incompressibility of $\hat{\vec{f}}$ in very little, if any, significant expressibility will be lost.}

Existing work~\cite{Kingravi2016,Whitman2017,Brian2019} have successfully used \emph{kernel functions} to represent flow fields in the past.
% In this paper, the idea of \emph{kernel embedding} is used to embed a set of flow vectors in a latent state through a kernel function.
% This allows modelling of a continuous flow field, where flow vectors at other positions can be queried. 
The \emph{incompressible kernel}~\cite{Brian2019} has recently been shown to be an apt description of smoothness and incompressibility of physical 2D flow fields.
The incompressible kernel~${\mat{K}:\realset^2\times\realset^2\to\realset^{2\times 2}}$ can be expressed as
\begin{equation} \label{eqn:incompressibleKernel}
    \mat{K}(\vec{x},\vec{x}^\prime) =
    \vec{\mathcal{D}}(\vec{x})k(\vec{x},\vec{x}^\prime)\transpose{\vec{\mathcal{D}}\left(\vec{x^\prime}\right)}
    ,
\end{equation}
where $\vec{\mathcal{D}}(\vec{x})=\transpose{\begin{bmatrix}\frac{\partial}{\partial y} & -\frac{\partial}{\partial x}\end{bmatrix}}$, and ${k:\realset^2\times\realset^2\to\realset}$ is an ``inner'' kernel function describing how the flow vectors are expected to be given their proximity.
Given prior knowledge of the underlying function, an appropriate inner kernel function can be chosen, which can lead to more accurate modelling and better convergence properties.

The Gram matrix~$\mat{K}(\mat{X},\mat{X}^\prime)$ is defined with $N_P$~positions~$\vec{x}_i$ as columns of~$\mat{X}\in\realset^{2\times N_P}$ and $N_Q$~positions~$\vec{x}^\prime_j$ as columns of~$\mat{X}^\prime\in\realset^{2\times N_Q}$:
\begin{equation} \label{eqn:incompressibleGrammian}
    % \mat{K}(\mat{X},\mat{X}^\prime) =  \begin{bmatrix}
    %     \mat{K}_{1,1}& \cdots & \mat{K}_{N_P,1} \\
    %     \vdots & \ddots & \vdots \\
    %     \mat{K}_{1,N_Q} & \cdots & \mat{K}_{N_P,N_Q} \\
    % \end{bmatrix}
    \mat{K}(\mat{X},\mat{X}^\prime) = \left[\mat{K}_{i,j}\right] \in \realset^{2N_Q\times2N_P}
    ,
\end{equation}
where $\mat{K}_{i,j} = \mat{K}(\vec{x}_i,\vec{x}^\prime_j)$.
Then, given ensemble data positions $\mat X_\text{ens}=[\vec{x}_1,\cdots,\vec{x}_{N_V}]$ and any spatial location~${\vec x\in\realset^2}$, the latent state vector $\beta\in\realset^{2N_V}$ encodes the flow vector $\hat{\vec f}(\vec x)$ at $\vec x$ 
via
% by assigning arguments $\mat X = \vec x, \mat X' = \mat X_\text{ens}$:
% An incompressible flow field~$\hat{\vec{f}}$ can then be modelled with a latent state~$\vec{\beta}\in\realset^{2N_V}$ as
\begin{equation} \label{eqn:flowPrediction_latent}
    \hat{\vec{f}}(\vec{x}) = \mat{K}(\vec{x},\mat{X}_\text{ens})\vec{\beta}
    .
\end{equation}
For fixed $\beta$, evaluating all $\vec x\in\realset^2$ using \eqref{eqn:flowPrediction_latent} results in a continuous 2D flow field $\hat{\vec f}(\vec x)$ that is incompressible. 
% where~$\vec x\in\realset^{2}$ is any spatial location, and \mat{X}_\text{ens}$ is the matrix form of positions~$\set{X}_\text{ens}$ such that~${\mat{X}_\text{ens}=[\vec{x}_1,\cdots,\vec{x}_{N_V}]}$.
% Then, the latent state~$\vec{\beta}_i$ of each ensemble member~$\vec{\eta}_i$ is obtained as the least squares solutions 
Then, for each ensemble member $\set e_i\in\set E$, the latent state $\beta_i$ is chosen by 
\begin{equation} \label{eqn:flowRegressionProblem}
    \beta_i = \argmin_{\vec{\beta}} \norm{\vec{\eta}_i- \mat{K}(\mat{X}_\text{ens},\mat{X}_\text{ens})\vec{\beta}}{2}^2
    ,
\end{equation}
where~$\vec{\eta}_i$ is the vectorised form of ensemble member~$\set{e}_i$, such that~${\vec{\eta}_i = \transpose{[u_{i,1},v_{i,1},\cdots,u_{i,N_V},v_{i,N_V}]}}$.

% \fk{Reviewer 6: I got a bit confused with the notation involving the
% arguments of incompressible kernel $\bf K$ in equation 6,
% equation 7 and also equation (11). I trust that $\mathbf x$ and
% $\mathbf X_{ens}$ have different dimensions...}

% The continuous nature of the latent state representation can be illustrated by querying the flow vectors at positions other than  $\mat{X}_\text{ens}$.
% Figure~\ref{fig:flowRegressioN_Example} shows an example ensemble member~$\vec{z}_i$ in black.
% The flow vectors in grey are queried using~\eqref{eqn:flowPrediction_latent}.
% \begin{figure}[tb]
%     \centering
%     \includegraphics[width=0.66\columnwidth]{flowRegressioN_Example1}
%     \caption{
%         Flow field regression of a single ensemble member~$\vec{f}_i$~(black) from an ensemble.
%         Flow vectors at other locations~(grey) can be computed through inference using~\eqref{eqn:flowPrediction_latent}.
%     }
%     \label{fig:flowRegressioN_Example}
% \end{figure}

In our approach, the estimated flow field is refined by adjusting values in the latent state representation.
Our choice of the kernel function $\mat{K}$ ensures that the resulting flow field is always incompressible.

\subsection{Model compression by SVD}
% The flow field model~$\hat{\vec{f}}_\vec{\beta}(\vec{x})$ can be compressed even further from the space of incompressible functions, because of the patterns that arise from the interaction with the environment~(e.g. topography). 
% We use a singular value decomposition~(SVD) to identify this lower-dimensional structure, and represent the model as a linear combination of at most $N_E$ \emph{basis flow fields}.
% These basis flow fields are recurring flow patterns that appears in the ensemble members at varying degrees.
% Since the dimensionality of the ensemble is typically much less than that of the variables in the latent state~($N_E \ll 2N_V$), this reduces the number of parameters needed to represent the flow field.
% Further compression can be achieved by excluding basis flow vectors that are insignificant in the reconstruction of the flow field.

In \eqref{eqn:flowPrediction_latent}, $\mat{K}$ could be considered a candidate $\mat{H}(\vec{x})$. 
However, it would not preserve the spatial correlations in $\set{E}$. 
This subsection addresses the construction of a $\mat{H}(\vec{x})$ so that the spatial correlations are preserved via the SVD. 
This admits a lower-dimensional representation of $\hat{\vec{f}}$ by additionally constraining $\hat{\vec{f}}$ at~$\set{X}_\text{ens}$ to be in the span of $\set E$, hence the name ``compression''.

First, the latent states of the ensemble members are concatenated for the \emph{latent matrix}~$\mat{B}\in\realset^{2N_V\times N_E}$
\begin{equation} \label{eqn:latentMatrix}
    \mat{B}=\begin{bmatrix}
        \vec{\beta}_1 & \vec{\beta}_2 & \cdots & \vec{\beta}_{N_E}
    \end{bmatrix}
    .
\end{equation}
Then the thin~SVD provides the compact representation:
\begin{equation} \label{eqn:decomposition}
    \mat{B}=\mat{U}\mat{\Sigma}\mat{V}^H
    ,
\end{equation}
where $(\cdot)^H$ is the conjugate transpose operator.
Since the number of ensembles is typically much less than that of the variables in the latent state~($N_E \ll 2N_V$), the left-singular vector~${\mat{U}\in\realset^{2N_V\times N_E}}$ is rectangular whilst the right-singular vector ${\mat{V}\in\realset^{N_E\times N_E}}$ and the diagonal matrix~${\mat{\Sigma}\in\realset^{N_E\times N_E}}$ of singular values~$\sigma_i$ are square.

The columns~of~$\mat{U}$ are referred to as the latent state components~$\vec{u}_i$, and corresponds to the latent states of the basis flow fields.
The singular values~$\sigma_i$ can be interpreted as the significance of the basis flow fields in the ensemble.
The columns~of~$\mat{V}$ describe the relative contribution of basis flow fields for the corresponding ensemble flow field.

The weight matrix~$\mat{W}\in\realset^{N_E\times N_E}$ is defined as
\begin{equation}
    \mat{W} = \mat{\Sigma}\mat{V}^H
    ,
\end{equation}
the columns of which are \emph{weight vectors}~$\vec{w}_i\in\realset^{N_E}$.
The basis~$\mat{H}$ can now be defined as
\begin{equation}
    \mat{H}(\vec{x}) = \mat{K}(\vec{x},\mat{X}_\text{ens})\mat{U}
    .
\end{equation}
% which is used with a weight vector~$\vec{w}$ to reconstruct the flow field in~\eqref{eqn:flowPrediction_weight}.
% \begin{equation} \label{eqn:flowPrediction_weight}
%     \hat{\vec{f}}_{\vec{w}}(\vec{x}) = \mat{H}(\vec{x})\vec{w}
%     .
% \end{equation}

\begin{figure}[tb]
    \centering
    \begin{subfigure}[b]{0.33\columnwidth}
        \includegraphics[width=\textwidth]{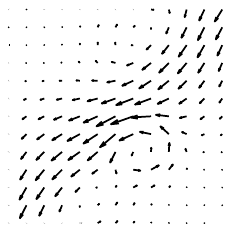}
        \preSubCaptionSpace{}
        \captionsetup{justification=centering}
        \caption{Ensemble member $\vec{\eta}_1$}
        \postSubCaptionSpace{}
        \label{subfig:ensemble_example}
    \end{subfigure}\hfill
    \begin{subfigure}[b]{0.66\columnwidth}
        \centering
        \includegraphics[width=\columnwidth]{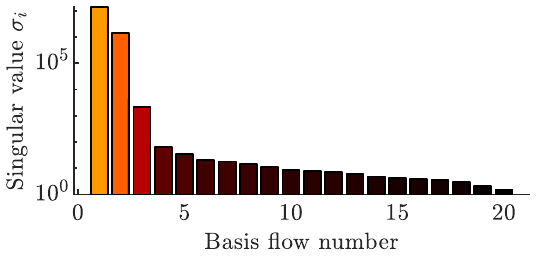}
        \preSubCaptionSpace{}
        \captionsetup{justification=centering}
        \caption{Basis flow singular values}
        \postSubCaptionSpace{}
        \label{subfig:singularValues_example}
    \end{subfigure}
    
    \begin{subfigure}[b]{0.33\columnwidth}
        \centering
        \includegraphics[width=\textwidth]{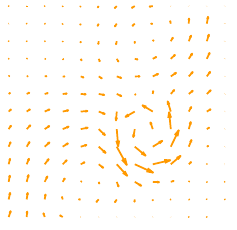}
        \preSubCaptionSpace{}
        \captionsetup{justification=centering}
        \caption{$\mat{K}(\mat{X}_\text{ens},\mat{X}_\text{ens})\vec{u}_1$}
        \postSubCaptionSpace{}
        \label{subfig:basisFlow_example1}
    \end{subfigure}\hfill
    \begin{subfigure}[b]{0.33\columnwidth}
        \centering
        \includegraphics[width=\textwidth]{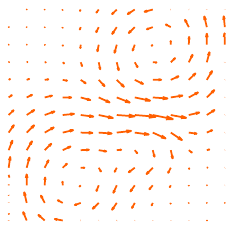}
        \preSubCaptionSpace{}
        \captionsetup{justification=centering}
        \caption{$\mat{K}(\mat{X}_\text{ens},\mat{X}_\text{ens})\vec{u}_2$}
        \postSubCaptionSpace{}
        \label{subfig:basisFlow_example2}
    \end{subfigure}\hfill
    \begin{subfigure}[b]{0.33\columnwidth}
        \centering
        \includegraphics[width=\textwidth]{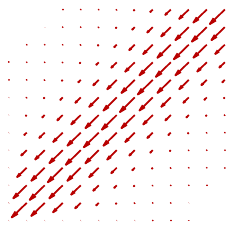}
        \preSubCaptionSpace{}
        \captionsetup{justification=centering}
        \caption{$\mat{K}(\mat{X}_\text{ens},\mat{X}_\text{ens})\vec{u}_3$}
        \postSubCaptionSpace{}
        \label{subfig:basisFlow_example3}
    \end{subfigure}
    \preCaptionSpace
    \caption{
        Model compression for ensemble flow field models.
        (\subref{subfig:singularValues_example}) shows the singular values in descending order.
        % The bars are coloured so that more significant singular values are brighter.
        % Note the vertical axis is plotted in log scale to emphasise the distinct difference in value.
        (\subref{subfig:basisFlow_example1}-\subref{subfig:basisFlow_example3}) shows the three most significant basis flow fields which can be used to reconstruct the flow field in (\subref{subfig:ensemble_example}).
    }
    \postCaptionSpace
    \label{fig:decomposition_example}
\end{figure}

The basis flow fields can be visualised by plotting~$\mat{H}(\vec{x})$ using only one latent state component~$\vec{u}_i$ instead of $\mat{U}$.
Figure~\ref{fig:decomposition_example} shows an example of model compression process using an ensemble consisting of flow fields similar to the one shown in Fig.~\ref{subfig:ensemble_example}.
The ensemble data example can largely be represented with the three basis flow fields~(\mbox{Fig.~\ref{fig:decomposition_example}\subref{subfig:basisFlow_example1}-\subref{subfig:basisFlow_example3}}), with the three largest singular values.

% >>> BORROWED SPACE
\begin{figure*}[tb]
    \centering
    \begin{subfigure}[b]{0.6\columnwidth}
        \centering
        \includegraphics[width=\textwidth]{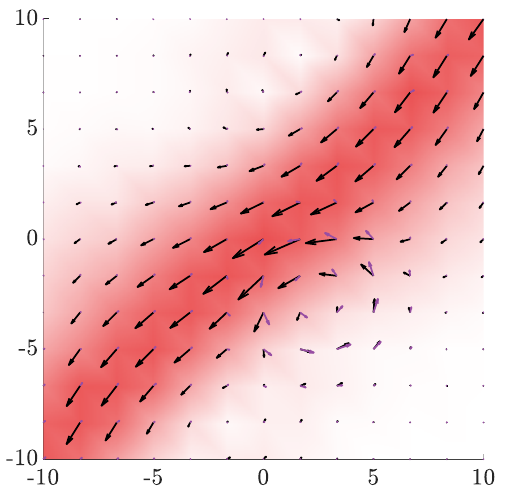}
        \preSubCaptionSpace{}
        \captionsetup{justification=centering}
        \caption{$k=0$}
        \postSubCaptionSpace{}
        \label{subfig:estimation_example0}
    \end{subfigure}\hfill
    \begin{subfigure}[b]{0.6\columnwidth}
        \centering
        \includegraphics[width=\textwidth]{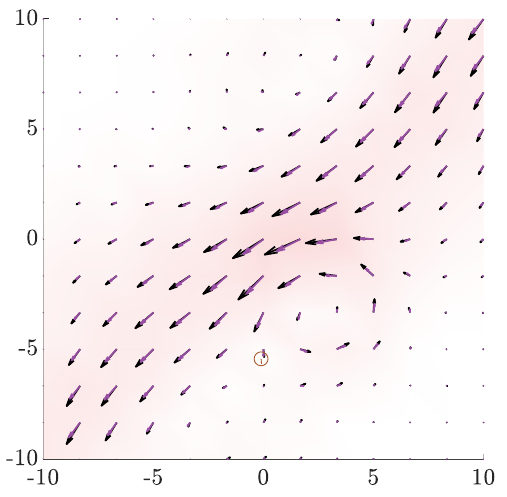}
        \preSubCaptionSpace{}
        \captionsetup{justification=centering}
        \caption{$k=1$}
        \postSubCaptionSpace{}
        \label{subfig:estimation_example1}
    \end{subfigure}\hfill
    \begin{subfigure}[b]{0.8\columnwidth}
        \centering
        \includegraphics[width=\textwidth]{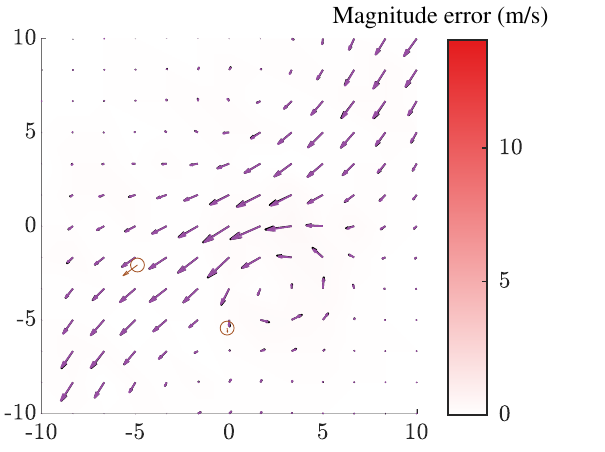}
        \preSubCaptionSpace{}
        \captionsetup{justification=centering}
        \caption{$k=2$}
        \postSubCaptionSpace{}
        \label{subfig:estimation_example2}
    \end{subfigure}
    \preCaptionSpace{}
    \caption{
        Estimated flow fields after $k$ measurements.
        The estimated flow field~(purple) quickly converges to the true flow field~(black) after a few measurements.
        The red heatmap shows the magnitude flow field error across space.
        Measurements are shown as circled brown arrows.
    }
    \postCaptionSpace{}
    \label{fig:estimation_example}
\end{figure*}
% <<<

% In our approach, the estimated flow field~$\hat{\vec{f}}$ is represented as a weight vector~$\hat{\vec{w}}$.
% Since this weight vector can be any column of~$\mat{W}$ and any value in between, the representation can be thought of being able to represent any deformation between the ensemble flow fields.

We can choose to represent the flow field with $N_W\le N_E$ weights by keeping only the highest $N_W$~singular values of the SVD and the corresponding columns~of~$\mat{U}$ and $\mat{V}$.
In particular, choosing $N_W<N_E$ effectively ignores the effects of basis flow fields that vary less in the ensemble.

% \fk{Say somewhere that choice of $\mat H$ means that we search only over the span of $\mat E$ (I think), which means the search space is much smaller. Convergence rate advantages are likely.}

% Note that the ensemble data~$\mat{F}$ should not be decomposed directly before applying regression on its modes, as the modes are guaranteed to be incompressible.

%-------------------------------------------------------------
% Approach: Online section
%-------------------------------------------------------------
\section{Online flow field estimation} \label{sec:online}
In the previous section, we described the offline stage of our approach, representing a flow field as a weight vector~$\vec{w}$ with each weight corresponding to the basis flow fields of the ensemble data.
In this section, we propose a process to incorporate information from noisy online measurements to improve estimation accuracy by recursively updating $\vec w$ given an initial estimate of the flow field.

Despite being static in this paper, we model the true flow field as a degenerate discrete-time linear dynamical system, which can easily be extended for use in the time-varying case. The process and measurement models of the true flow field are
\begin{align}
    \vec{w}_k &= \mat{A}\vec{w}_{k-1} \label{eqn:stateTransition}\\
    \vec{z}_k &= \mat{H}(\vec{x}_k)\vec{w}_k + \vec{n}_k \label{eqn:measurementModel}
    ,
\end{align}
where $\mat{A}=\mat{I}_{N_W\times N_W}$ is the state transition model for the static flow field, and ${\vec{n}_k\sim\mathcal{N}(\vec{0}_{2\times 1},\mat{\Sigma}_\text{mea})}$ is the i.i.d. measurement error with covariance~$\mat{\Sigma}_\text{mea}$.
The Kalman filter~\cite{Kalman1960} is employed as it is the optimal estimator for measurements with Gaussian noise~\cite{Humpherys2012}; the Kalman filter equations are well-known (e.g. \cite{Thrun2005}) and are omitted for brevity.

As a recursive estimator, the Kalman filter needs to be initialised with some estimate.
The columns of the weight matrix~$\mat{W}$ are the weight vectors that correspond to each ensemble flow field, so an estimate can be obtained by aggregating them.
An initial estimate weight vector~$\hat{\vec{w}}_0$ and its covariance~${\mat{P}_0\in\realset^{N_W\times N_W}}$ is defined as the row-wise mean and the diagonal matrix of row-wise variances of the weight matrix~$\mat{W}$.

% To refine this weight estimate, notice that our flow field representation~\eqref{eqn:flowPrediction_weight} of the static flow field~$\vec{f}_\text{tru}$ can be closely approximated as a ``degenerate'' linear dynamical system with respect to the values of $\vec{w}$.

% $\vec w^\star$, the ``least squares weight vector''

% Intuitively, the Kalman filter compares measurements of the flow field with the model's predicted value and adjusts the weight vector based on the relative uncertainty of the estimate and the measurement.
% Since the spatial correlations in~$\mat{H}(\vec{x})$ describe flow vectors across the entire space, the estimator makes global corrections instead of flow vectors in the local area.
% By refining weights instead of the latent state, the estimator makes global corrections of the flow field instead of flow vectors in the local area.

% By refining weights instead of the latent state, we focus estimation efforts to areas of the flow field that are more uncertain in the ensemble data.

Figure~\ref{fig:estimation_example} shows an example of the estimated flow fields after $k$~measurements.
The rapid convergence to the true flow field can be attributed to a good initial estimate from the ensemble data, and well-placed measurements.

%% Kalman filter update equations
% \begin{align}
%     \hat{\vec{z}}_k &= \mat{H}_k\hat{\vec{w}}_{k|k-1} \\
%     \tilde{\vec{y}}_k &= \vec{z}_k - \hat{\vec{z}}_k \\
%     \mat{S}_k &= \mat{H}_k\mat{P}_{k|k-1}\transpose{\mat{H}_k} + \mat{R}_k \\
%     \mat{K}_k &= \mat{P}_{k|k-1}\transpose{\mat{H}_k}\mat{S}_k^{-1} \\
%     \hat{\vec{w}}_{k|k} &= \hat{\vec{w}}_{k|k-1} + \mat{K}_k\tilde{\vec{y}}_k \\
%     \mat{P}_{k|k} &= \left(\mat{I}_{w\times w}-\mat{K}_k\mat{H}_k\right)\mat{P}_{k|k-1}
% \end{align}

%-------------------------------------------------------------
% Discussion/Analysis
%-------------------------------------------------------------
\section{Analysis} \label{sec:analysis}

% >>> BORROWED SPACE
\begin{figure*}[tb]
    \centering
    \begin{subfigure}[b]{49.67mm}
        \centering
        \includegraphics[width=\textwidth]{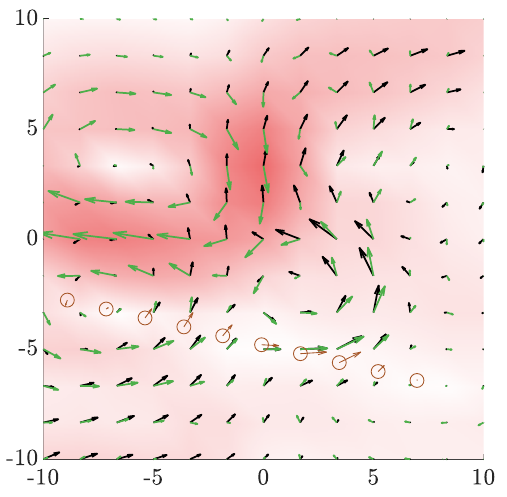}
        \preSubCaptionSpace{}
        \captionsetup{justification=centering}
        \caption{Kernel observer~\cite{Kingravi2016} estimate}
        \postSubCaptionSpace
        \label{subfig:simulated_flow_ko}
    \end{subfigure}\hfill
    \begin{subfigure}[b]{49.67mm}
        \centering
        \includegraphics[width=\textwidth]{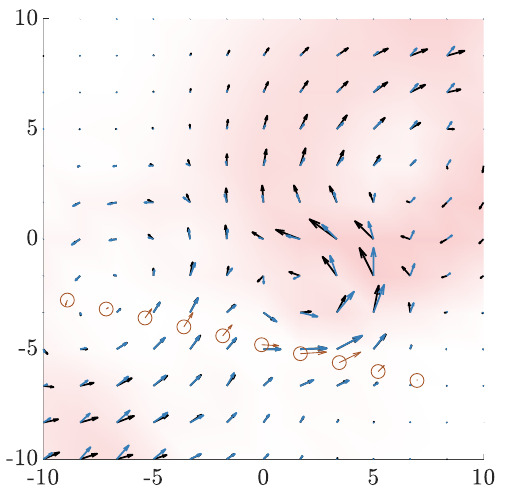}
        \preSubCaptionSpace{}
        \captionsetup{justification=centering}
        \caption{Incompressible GP~\cite{Brian2019} estimate}
        \postSubCaptionSpace
        \label{subfig:simulated_flow_gp}
    \end{subfigure}\hfill
    \begin{subfigure}[b]{64.66mm}
        \centering
        \includegraphics[width=\textwidth]{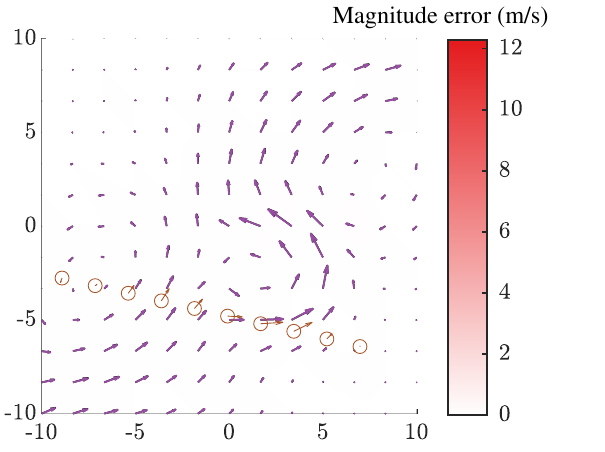}
        \preSubCaptionSpace{}
        \captionsetup{justification=centering}
        \caption{Our approach}
        \postSubCaptionSpace
        \label{subfig:simulated_flow_kf}
    \end{subfigure}
    \preCaptionSpace{}
    \caption{
        Comparison of flow field estimation using equally-spaced measurements~(circled brown arrows) from the bottom right to the left.
        The magnitude error between the estimated and the true flow field~(black) is shown as a red heatmap.
    }
    \postCaptionSpace{}
    \label{fig:simulated_flow}
\end{figure*}
% <<<

% estimation accuracy and representation discussion
% - In order to evaluate the statistical characteristics at each ensemble point~$\mathbf{x} \in \mathbf{X}_{\text{ens}}$, we use all~$N_V \cdot N_E$ flow vectors while the standard uses only $N_E$ flows; only the~$1/N_V$ proportion of data is used to estimate. Since the number of ensemble points in forecast can be quite large depending on the size of environment, it significantly influences the overall estimation performance.

% computational complexity
% - We are dealing with $N_V \cdot N_E$ individual flow vectors in~$E$ to find statistical characteristics at each point while the standard uses~$N_E$
% - Overall, the computational complexity for standard method is~$N_V \cdot N_E$ for all ensemble points

Assuming that the environment is large, the number of ensemble positions~$N_V$ is greater than the number of ensemble members in~$\set{E}$ (i.e. $N_E \ll N_V$).
Under this assumption, we demonstrate the worst-case computation time complexity in Table~\ref{table:complexity} for our framework as well as for kernel observer~(KO)~\cite{Kingravi2016}, Gaussian process~(GP)~\cite{Brian2019}, and least squares regression~(LS) methods.

\begin{table}[tb]
    \centering
    \caption{
        Computational complexity of compared approaches
        % , our method with least squares~(LS), kernel observer~(KO), and incompressible GP~(GP)
        % Worst-time computation complexities shown for initialisation, update, and query.
        % Our method is compared against three approaches: kernel observer (KO), incompressible GP (GP) and our method with least squares (LS).
    }
    \vspace{-0.5em}
    \label{table:complexity}
    \begin{tabular}{r|c c c}
        & Initialise & Update & Query \\
        \hline
        KO & $\bigOh{N_V^3}$ & $\bigOh{N_V^3}$ & $\bigOh{N_V}$ \\
        GP & $\bigOh{N_V}$ & $\bigOh{1}$ & $\bigOh{(N_K + N_V)^3}$ \\
        LS & $\bigOh{N_V^3}$ & $\bigOh{N_W^3+N_K N_W^2}$ & $\bigOh{N_V N_W}$ \\
        Ours & $\bigOh{N_V^3}$ & $\bigOh{N_W^3+N_V N_W}$ & $\bigOh{N_V N_W}$
    \end{tabular}
    \vspace{-2em}
\end{table}

The kernel observer approach is proposed to estimate a time-varying latent state with very few measurements.
For comparison, in the time-invariant case with ensembles, we adapt the framework by setting the transition operator to the identity, and initialising its estimate with the ensemble latent states.
This adaptation is identical to our approach up to forming the latent matrix in~\eqref{eqn:latentMatrix}, after which the mean and variance of $\mat{B}$ is used to initialise their Kalman filter.
% in~\eqref{eqn:decomposition} in which we find the weight matrix~$\mathbf{W}$ that reduces the amount of data required to reconstruct the ensemble~$\mathbf{E}$.
The GP approach utilises a kernel that allows for an estimate of incompressible flow fields from flow vector measurements and is adapted for use with ensembles by finding the set of flow vector mean and covariances over individual ensemble positions~$\set{X}_\text{ens}$, which is then added as prior flow field measurements.
Whilst the Kalman filter shares the same estimation performance as least squares, a least squares variation of our approach is included to highlight the computational disadvantages in the context of our problem.
A useful aspect of the least squares approach is that it does not need to be initialised with an estimate.
Note that all the listed approaches generate incompressible flow fields through the use of the incompressible kernel~\cite{Brian2019}.

In Table~\ref{table:complexity}, we have time complexities for three operations: 1) \emph{initialise}~(with ensemble forecast), 2) \emph{update} where the estimate~$\hat{\vec{f}}$ is updated with a single measurement, and 3) \emph{query} where a flow vector estimate is queried at an arbitrary position.
The complexities are described with the number of ensemble positions~$N_V$, the number of weights~$N_W$, and the number of accumulated measurements~$N_K$.

For those methods with similar steps to our method~(i.e. KO, LS), the time complexity is~$\bigOh{N_V^3}$, whereas that for the GP is~$\bigOh{N_V}$.
This is because the GP computes the flow field lazily, and just stores the $N_V$~aggregated flow vectors from the ensemble.
Ours computes the latent representation of each ensemble flow field for basis flow field extraction.
The basis flow fields captures the spatial correlations of flow vectors in the ensemble data.
Unlike our method, the GP loses significant amount of information since it is unable to capture this spatial correlation.

The complexity for update is also similar among the KO, LS and our method with two important differences.
Our method reduces the number of estimated variables by compressing a representation with~$\vec{\beta}$ to the low-dimensional representation~$\vec{w}$, unlike the KO approach where compression is omitted.
Since~$N_W\leq N_E \ll N_V$, ours outperforms the KO even without truncation~(i.e. when $N_W = N_E$).
For the LS method, one particular implementation is to update the weight vector after each measurement, which requires all $N_K$~measurements to be considered.
This particular implementation becomes intractable as the number of past measurements grows large over time.
For the GP, the update is performed by concatenating the new measurement to the existing set.

For querying a position, the GP method requires an inversion of the Gram matrix~\eqref{eqn:incompressibleGrammian} between the positions of the collected measurements and itself.
This is an expensive operation that is cubic in complexity~\cite{Rasmussen2006}, so as the number of measurements becomes large, the time complexity approaches~$\bigOh{N_K^3}$.
% In closed-loop path planning where we keep plan, execute and measure, the GP method dramatically slows down the overall planning time as the mission duration gets longer.

As a whole, our method directly exploits the incompressible basis flow fields present in the ensemble members to fit the weights to the measurements.
The compact representation of flow fields as weights significantly reduce the time complexities for update and query operations with compression and truncation.
We show that these operations are not dependent on the number of past measurements.
This is an important property, especially in long-term missions and  closed-loop path planning where a robot continuously measures the flow field to update its estimate.
\section{Empirical results}
% In this section, our proposed framework is demonstrated in two simulated scenarios: one scenario involving synthetic flow fields, and another with real forecast data from the Australian Bureau Meteorology~(BOM).
% In the first scenario, we compare the performances of two existing approaches, our approach, and our approach with a least squares~(LS) implementation with a single vehicle taking measurements.
% The second scenario uses real ensemble data from the Australian BOM to demonstrates how the performance of our approach changes with different measurement policies, possibly for a network of sensors or from a team of coordinated robots.

In this section, we compare the performance of our method against existing approaches.
We show that our method reduces the estimation error across the environment while the existing approaches are limited to the neighbourhood of measurements. We also empirically validate the theoretical properties for update and query time. 
Then, we use real ensemble forecast provided by the BOM to discuss how different measurement policies affect the accuracy of flow estimation.
We argue how our method could be implemented for various planning problems, especially \emph{active perception}~\cite{Fred2019,Brian2018,Bajcsy2018}, where regular updates with measurements is necessary.
In all demonstrations, we use the incompressible kernel in~\eqref{eqn:incompressibleKernel} with the squared exponential kernel~\cite{Rasmussen2006}:
\begin{equation}
    k(\vec{x},\vec{x}^\prime) = k_\text{SE}(\vec{x},\vec{x}^\prime) = \sigma_\text{ker}^2 \exp{\left(\frac{\norm{\vec{x}^\prime-\vec{x}}{2}^2}{2\ell^2}\right)}
    ,
\end{equation}
where $\ell$ and $\sigma_\text{ker}$ are hyperparameters to be tuned to the flow field of interest.

% In all demonstrations, we use the squared-exponential kernel~${k_\text{SE}:\realset^2\times\realset^2\to\realset}$ as the inner kernel function of~\eqref{eqn:incompressibleKernel}:

% \subsection{Comparison with existing work in synthetic flow}
\subsection{Estimation comparison with existing approaches}

We compare our method against the incompressible GP (GP)~\cite{Brian2019}, an adaptation of the kernel observer~\cite{Kingravi2016} with Kalman filter (KO), and the least squares implementation of our method (LS).
We synthetically construct an ensemble of $N_E=20$~flow fields over $N_V=169$ positions, shown in Fig.~\ref{fig:simulated_flow}.
The true flow field~$\vec{f}_\text{tru}$ is shown with black arrows.
Starting at $\vec{x}=\transpose{[6.98,-6.42]}$, the vehicle makes a set of $10$~measurements with noise~${\mat{\Sigma}_\text{mea}=10^{-3}\mat{I}_{2\times 2}}$ that are $2$\,\si{m} apart while moving in a straight line.
The measurement positions are shown with brown circles with measurement vectors as arrows.

% We constructed the true flow field and $20$~ensemble member flow fields via the superposition of three component flow fields, similar to the ones in \mbox{Fig.~\ref{fig:decomposition_example}\subref{subfig:basisFlow_example1}-\subref{subfig:basisFlow_example3}} with various magnitudes of flow speed.
% The true flow field~$\vec{f}_\text{tru}$ over $N_V=169$~positions (black arrows) in Fig.~\ref{fig:simulated_flow} has a maximum flow speed of $5.92$\,\si{m/s}, and is not included as prior knowledge for any of the approaches.
% The ensemble data uses a square grid of $N_V=169$~positions.
% The vehicle makes measurements around $2$\,\si{m} apart with noise~${\mat{\Sigma}_\text{mea}=\tilde{\mat{\Sigma}}_\text{mea}=10^{-3}\mat{I}_{2\times 2}}$ in a straight line from the bottom right to the left.
% All approaches use the same measurement covariance~$\tilde{\sigma}_\text{mea}$, and the incompressible kernel as described in~\eqref{eqn:incompressibleKernel} with hyperparameters $\ell=3$\,\si{m} and $\sigma_\text{ker}=8.46$ tuned by hand for this scenario.

In Fig.~\ref{subfig:simulated_flow_ko}, the flow field estimate from KO is shown.
The KO method utilises the latent state of the kernel~$\vec{\beta}$ with the size of $338$~(i.e. $2N_V$).
We observe large errors; we suspect that due to the large latent state, the problem is still underdetermined.

In Fig.~\ref{subfig:simulated_flow_gp}, we show the estimates using incompressible GP~\cite{Brian2019} over the synthetic ensemble data.
The estimation after the measurements is shown with blue arrows and the error magnitude is shown in red.
As discussed in Sec.~\ref{sec:analysis}, the GP method is unable to make use of all the spatial correlations present in the ensemble resulting in lower sampling efficiency.
As a consequence, the error is only reduced in the neighbourhood of measurements while the distant regions are virtually not affected.

The estimates from our method are shown in Fig.~\ref{subfig:simulated_flow_kf}. 
Using our method, decomposition in Sec.~\ref{sec:offline} found three representative basis flow fields.
As a consequence, the number of basis flows can be reduced from $20$~(i.e. $N_E$) to $3$~(i.e. $N_W$) after truncation with virtually no degradation in estimation performance.
We show that our method significantly reduces distant errors, unlike the GP and KO methods.
% This is because our method updates the estimates based on a set of most representative basis flows in smaller size.
Unlike the KO method, the number of unknown variables in our approach is also much smaller (i.e. $N_W \ll 2N_V$) so less measurements are required to properly determine the variables.
The results also show that the distant positions from measurements are well estimated since our method considers spatial correlations in flow fields unlike the GP method.

% Our approach (LS+KF) and its flow estimate
% For our approach, the decomposition leads to three basis flow fields that are very similar to the ones used to generate the synthetic data, so we can afford to truncated the number of weights to ${N_W=3}$.
% This allows the estimate to quickly converge globally to the true flow field with very few measurements.
% Both LS and our proposed implementations perform well and lead to visually identical flow fields so only the estimate from our approach is shown in Fig~\ref{subfig:simulated_flow_kf}.

\begin{figure}[tb]
    \centering
    \begin{subfigure}[b]{0.5\columnwidth}
        \centering
        \includegraphics[width=\textwidth]{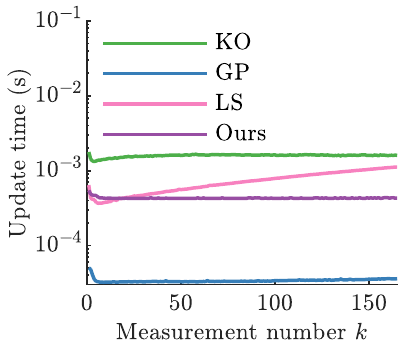}
        \preSubCaptionSpace
        \captionsetup{justification=centering}
        \caption{Iteration time per measurement}
        \postSubCaptionSpace
        \label{subfig:iterComp}
    \end{subfigure}\hfill
    \begin{subfigure}[b]{0.5\columnwidth}
        \centering
        \includegraphics[width=\textwidth]{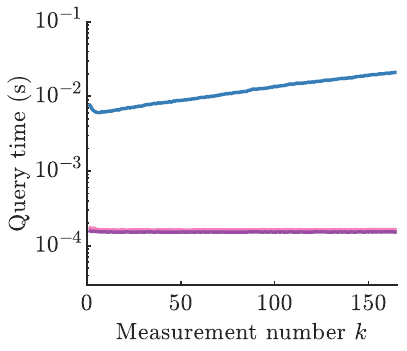}
        \preSubCaptionSpace
        \captionsetup{justification=centering}
        \caption{Query time per position}
        \postSubCaptionSpace
        \label{subfig:queryComp}
    \end{subfigure}
    \preCaptionSpace{}
    \caption{
        Comparison of mean computational times between the kernel observer~(KO)~\cite{Kingravi2016}, incompressible Gaussian process~(GP)~\cite{Brian2019}, a least squares~(LS) implementation of our approach, and our approach on the flow field shown in Fig.~\ref{fig:simulated_flow}.
        Values are averaged over $1000$ trials, the $99.73\%$ confidence intervals of which are virtually imperceptible, and are omitted.
    }
    \postCaptionSpace{}
    \label{fig:durationComparison}
\end{figure}

In Fig.~\ref{fig:durationComparison}, we show how our method scales in the number of past measurements~$N_K$.
As claimed in Sec.~\ref{sec:analysis}, the update time shown in Fig.~\ref{subfig:iterComp} stays almost constant in the number of measurements except the LS method.
It is important to note that the update time for our method is much lower than the KO due to our compressed representation (i.e. $N_W \ll 2N_V$).
Likewise, the query time shown in Fig.~\ref{subfig:queryComp} stays constant for our method while it increases for the GP method which is shown to be~$\bigOh{N_K^3}$ in Sec.~\ref{sec:analysis}.
Overall, the empirical comparison shows that our method scales in a suitable way for applications that take a large number of measurements.

% % Further comparison of computation times
% Decreasing the distance between measurements, we can compare computational times between the approaches along with the LS implementation of our approach.
% Figure~\ref{subfig:iterComp} shows the computation time needed for each approach for each measurement.
% The GP approach requires very little constant computational time as it simply needs to store the measurement.
% The KO approach uses more constant computation time due to the inversion of larger matrices involved in a Kalman filter with $2N_V=338$ variables.
% Our approach also requires very little constant computation time due to a smaller state vector of size $N_W=3$.
% The LS implementation of our approach uses an increasing amount of computation time due to the increasingly large matrices to solve for estimated weight vector \ct{maybe mention the computation complexity}.
% An alternate might involve computing the weight vector only when a query is needed, however the computation time will still be increasing for every measurement.

% Figure~\ref{subfig:queryComp} shows the computation time needed for a flow velocity query at a single position for each approach.
% The GP approach requires increasing amount of computation time for a query, due to the increasing number of measurements it gathers \ct{maybe mention the computation complexity}.
% The other approaches require very little constant computation since they all involve simple matrix multiplications.

\subsection{Measurement policies in real ensemble flow fields}

We show how different measurement policies affect the estimation performance using our method.
We use the ensemble forecast data provided by the BOM where a portion of the data is shown in Fig.~\ref{fig:bom_ensemble}.
We compare three policies: \emph{uniform} where measurement positions are randomly taken, \emph{subspace} where measurements are only taken within a subsection with high uncertainty, and \emph{active} where measurements are taken at positions with flow vector high uncertainty.

\begin{figure}[tb]
    \centering
    \includegraphics[width=\columnwidth]{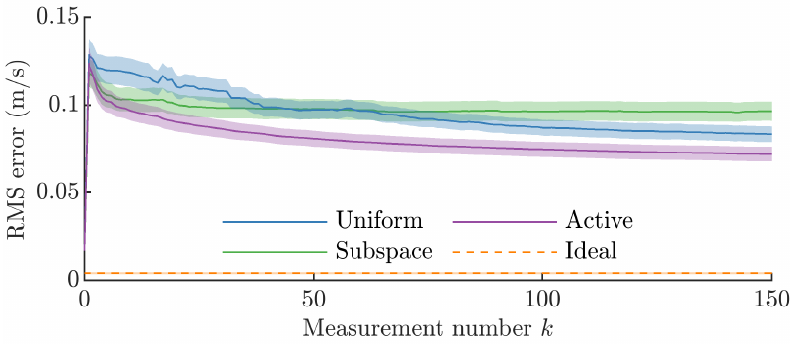}
    \preCaptionSpace{}
    \caption{
        Root mean square~(RMS) error from flow field estimation using different measurement policies using our approach.
        The flow field is based on the real ensemble data shown in Fig.~\ref{fig:bom_ensemble}.
        % The LS solution to noise-free measurements at $\mat{X}_\text{ens}$ is marked as ``ideal''.
        % Estimation from a sequence of measurements at uniformly random positions is marked as ``uniform''.
        % Estimation from a sequence of measurements at uniformly random positions constrained to the bottom left quadrant of the space is marked as ``subspace''.
        % Estimation from a sequence of measurements iteratively chosen at positions with high flow vector uncertainty is marked as ``active''.
        The $99.73\%$ confidence intervals for these errors are shown for each measurement policy.
    }
    \postCaptionSpace{}
    % TODO:
    % - Add process noise?
    \label{fig:bom_scenario}
\end{figure}

In Fig.~\ref{fig:bom_scenario}, we show the root mean square~(RMS) error with increasing number of past measurements.
For reference, we have the \emph{ideal} condition where measurements are taken exactly at ensemble positions~$\set{X}_\text{ens}$ such that the the final error after all measurements is shown with dashed orange line. 
Since ground-truth is not known for this forecast data, the error for each policy is evaluated using \emph{leave-one-out cross-validation}~(LOOCV)~\cite{Bishop2006}, where we initialise the estimates without one chosen ensemble member and compare the estimates against that chosen member.
We then compute the RMS of the residuals at~$\set{X}_\text{ens}$.
For each measurement policy, a trial is performed once with a fixed random seed.
The RMS error is shown in coloured lines and its~$3\sigma$ confidence interval is shown as a shaded band.

% Due to the more complex nature of the flow field, no truncation is performed for our approach, i.e. $N_W=N_E$, furthermore we standardise the latent states before performing SVD in~\eqref{eqn:decomposition}.
% The hyperparameters are $\ell=32.5$\,\si{km} and $\sigma_\text{ker}=11950$ tuned by hand for this scenario.

% The performance of each measurement pattern is measured as the normalised LS error, i.e. the LS error divided by number of elements in $\hat{\vec{f}}(\mat{X}_\text{ens})$.
% For reference, the ideal flow field estimate obtained as the LS solution with $\vec{f}_\text{tru}(\mat{X}_\text{ens})$ is referred to as ``ideal''.
% The errors across the $96$ tests for each measurement are summarised in Fig.~\ref{fig:bom_scenario}.

The results show that the active policy outperforms all other policies.
It is interesting to observe that the convergence rate to the ideal is much faster using our approach.
The uniform policy initially underperforms compared to the subspace because the subspace policy focuses the measurements in a region with high uncertainty so its reduction in error is larger.
However, as the number of measurements increases, the uniform policy starts to outperform the subspace policy.
This is because the subspace policy is constrained to a small region whereas the uniform policy takes measurements across the environment.

The policy comparison result indicates that existing path planning approaches will benefit from our method in that a carefully-chosen measurement policy will greatly affect the quality of the estimated flow field which will in turn affect the quality of the planned path. 
For example, we can formulate an active perception problem that involves finding a set of unknown features over uncertain ocean currents.
Unlike the traditional problems, we do not simply maximise our information over the features but also over ocean currents since practical ocean vehicles are typically advected by ocean currents~\cite{Inanc2005,Claus2010,Wang2016}.
Our method is both theoretically and empirically validated for use in such a problem setting.

\section{Conclusion and future work}
%%%%%%%%%%%%%%
% Conclusion
% 
% (In what way we have achieved the objectives set out in the paper)
% (Kind of elevator pitchy)
% We have presented an algorithm for extracting basis flow fields from ensemble forecasts using a combination of an incompressible kernel and SVD.
In this paper we have presented a novel algorithm for ocean flow field estimation. By leveraging ensemble data as a prior and integrating sequential online measurements, fast convergence to a good estimate is achieved in synthetic and real-world conditions.

%A Kalman filter is used to estimate the contributions of each basis flow field for the estimated flow field.

One limitation is the assumption of time-invariant flows, but ensemble data is available over large time windows.
Blending dynamic mode decomposition~(DMD) approaches for the corresponding approximate linear dynamical system could draw out the full potential of our method in the time-varying case.
% enable extractions of flow patterns across the dimension of time as well as the dimension of possibility.
An interesting immediate application is to exploit the spatially-correlated probabilistic representation of the flow field in path planning~\cite{Cadmus2019ICRA,Cadmus2019CDC,Cadmus2020,Chanyeol2021,Chanyeol2019,Giovanni2021} in uncertain flows.

%%%%%%%%%%%%%%
% Extension of work (something we immediately plan to do)
% 
% Can do SLAM-like framework
% Ocean current estimation and navigation (with given cost function)
% Active perception
% Possible use of UKF or PF
% K-SVD for sparse modelling (L1 minimisation instead of L2) (compressed communications)
% ``Sparse GP'': Additional spatial compression of data (potentially more compact representation)
% - Since each representative point carries more responsibility in flow field representation, it might lead to more meaningful basis vectors after decomposition
% \fk{Consider the Ekman divergence, which is a real-world 3D phenomenon introduces slight violations of incompressibility when considering a 2D flow field.}
% Usage of different kernel hyperparameters for each basis flow field
% Can potentially incorporate various types of measurements
% - Curl
% - From drift
% - Stream value from one point to another, using vehicle velocity and displacement
% 
%%%%%%%%%%%%%%
% Applications in other problems
% 
% Planning in uncertain flow fields? \cite{Dey2014}
% Incompressible representation enables use of streamline-based planning

%-------------------------------------------------------------
% References
%-------------------------------------------------------------
\balance
\bibliographystyle{IEEEtran}
\bibliography{IEEEabrv,ref}

\end{document}